\documentclass[sigconf]{acmart}

\usepackage{booktabs} 
\usepackage{threeparttable}

\usepackage{mathrsfs}
\usepackage{bm}

\setcopyright{rightsretained}

\begin{document}
\title{Entire Space Multi-Task Model: An Effective Approach for Estimating Post-Click Conversion Rate }


\author{
Xiao Ma, Liqin Zhao, Guan Huang, Zhi Wang, Zelin Hu, Xiaoqiang Zhu, Kun Gai \\
Alibaba Inc. \\
 \{maxiao.mx, liqin.zlq, bingyi.wz, xiaoqiang.zxq, jingshi.gk\}@alibaba-inc.com }

\begin{abstract}
Estimating post-click conversion rate (CVR) accurately is crucial for ranking systems in industrial applications such as recommendation and advertising. Conventional CVR modeling applies popular deep learning methods and achieves state-of-the-art performance. However it encounters several task-specific problems in practice, making CVR modeling challenging. For example, conventional CVR models are trained with samples of clicked impressions while utilized to make inference on the entire space with samples of all impressions. This causes a \textit{sample selection bias} problem. Besides, there exists an extreme \textit{data sparsity} problem, making the model fitting rather difficult. In this paper, we model CVR in a brand-new perspective by making good use of sequential pattern of user actions, i.e., $impression \rightarrow click \rightarrow conversion$. The proposed \textbf{E}ntire \textbf{S}pace \textbf{M}ulti-task \textbf{M}odel (ESMM) can eliminate the two problems simultaneously by i) modeling CVR directly over the entire space, ii) employing a feature representation transfer learning strategy. Experiments on dataset gathered from traffic logs of Taobao's recommender system demonstrate that ESMM significantly outperforms competitive methods. We also release a sampling version of this dataset to enable future research. To the best of our knowledge, this is the first public dataset which contains samples with sequential dependence of click and conversion labels for CVR modeling.
\end{abstract}


\maketitle

\section{Introduction}

Conversion rate (CVR) prediction is an essential task for ranking system in industrial applications, such as online advertising and recommendation etc. 
For example, predicted CVR is used in OCPC (optimized cost-per-click) advertising to adjust bid price per click to achieve a win-win of both platform and advertisers~\cite{zhu2017optimized}. It is also an important factor in recommender systems to balance users' click preference and purchase preference.

In this paper, we focus on the task of post-click CVR estimation. To simplify the discussion, we take the CVR modeling in recommender system in e-commerce site as an example. Given recommended items, users might click interested ones and further buy some of them. In other words, user actions follow a sequential pattern of $impression \rightarrow click \rightarrow conversion$. In this way, CVR modeling refers to the task of estimating the post-click conversion rate, i.e., $\textit{pCVR} = p(conversion|click,impression)$.

\begin{figure}[!t]
\centering
\includegraphics[height=32mm ]{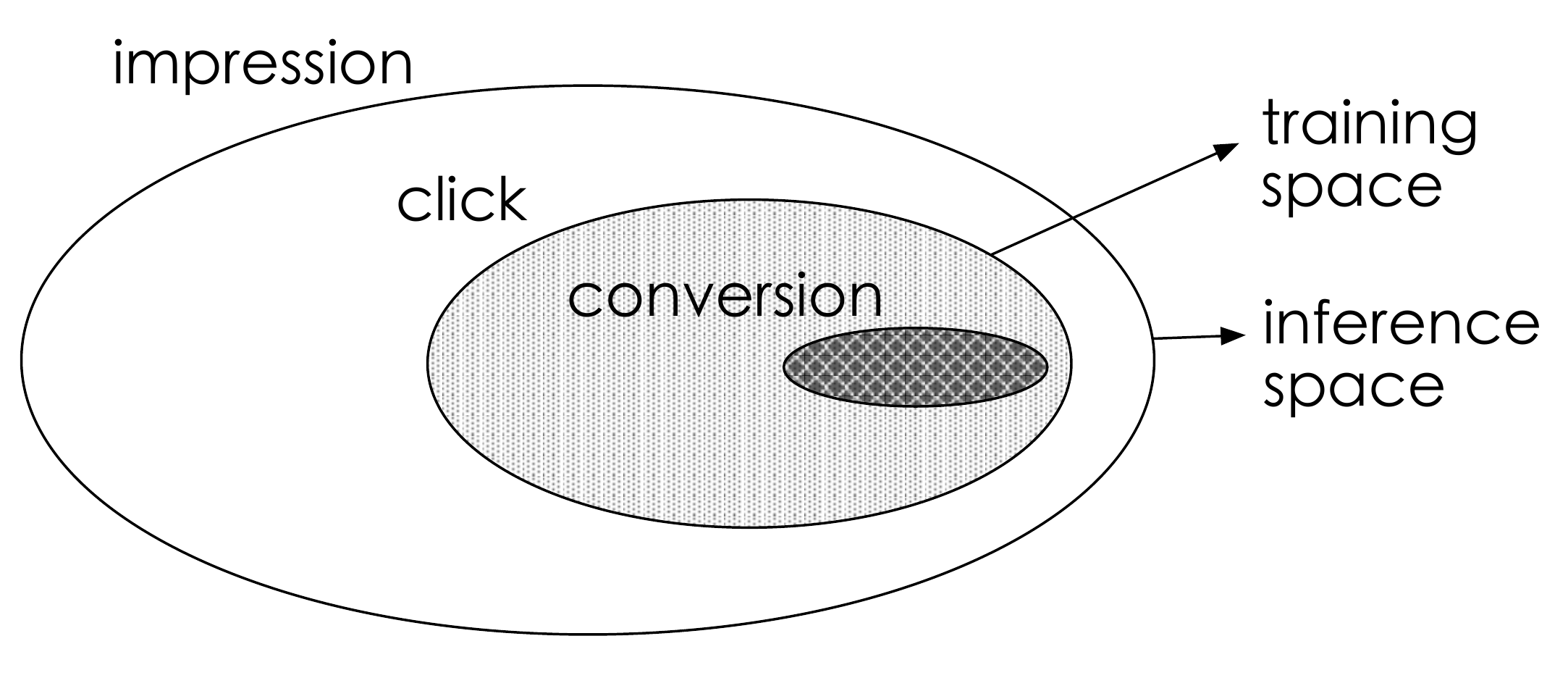}
\caption{{Illustration of sample selection bias problem in conventional CVR modeling. Training space is composed of samples with clicked impressions. It is only part of the inference space which is composed of all impressions.}}
\label{fig:pl}
\end{figure}

In general, conventional CVR modeling methods employ similar techniques developed in click-through rate (CTR) prediction task, for example, recently popular deep networks~\cite{widedeep,din}. However, there exist several task-specific problems, making CVR modeling challenging. Among them, we report two critical ones encountered in our real practice: 
i) \textsl{sample selection bias (\textbf{SSB})} problem~\cite{zadrozny2004learning}. As illustrated in Fig.\ref{fig:pl}, conventional CVR models are trained on dataset composed of clicked impressions, while are utilized to make inference on the entire space with samples of all impressions. 
\textsl{SSB} problem will hurt the generalization performance of trained models. 
ii) \textsl{data sparsity (\textbf{DS})} problem. In practice, data gathered for training CVR model is generally much less than CTR task. Sparsity of training data makes CVR model fitting rather difficult.

There are several studies trying to tackle these challenges. In \cite{ecvr}, hierarchical estimators on different features are built and combined with a logistic regression model to solve \textsl{DS} problem. 
However, it relies on a priori knowledge to construct hierarchical structures, which is difficult to be applied in recommender systems with tens of millions of users and items.
Oversampling method~\cite{weiss2004mining} copies rare class examples which helps lighten sparsity of data but is sensitive to sampling rates. All Missing As Negative (AMAN) applies random sampling strategy to select un-clicked impressions as negative examples ~\cite{pan2008one}.  It can eliminate the \textsl{SSB} problem to some degree by introducing unobserved examples, but results in a consistently underestimated prediction.  
Unbiased method~\cite{zhang2016bid} addresses \textsl{SSB} problem in CTR modeling by fitting the truly underlying distribution from observations via rejection sampling. However, it might encounter numerical instability when weighting samples by division of rejection probability. 
In all, neither \textsl{SSB} nor \textsl{DS} problem has been well addressed in the scenario of CVR modeling, and none of above methods exploits the information of sequential actions. 

In this paper, by making good use of sequential pattern of user actions, we propose a novel approach named Entire Space Multi-task Model (ESMM), which is able to eliminate the \textsl{SSB} and \textsl{DS} problems simultaneously.
In ESMM, two auxiliary tasks of predicting the post-view click-through rate (CTR) and post-view click-through\&conversion rate (CTCVR) are introduced. 
Instead of training CVR model directly with samples of clicked impressions, ESMM treats \textit{pCVR} as an intermediate variable which multiplied by \textit{pCTR} equals to \textit{pCTCVR}.  
Both \textit{pCTCVR} and \textit{pCTR} are estimated over the entire space with samples of all impressions, thus the derived \textit{pCVR} is also applicable over the entire space. 
It indicates that \textsl{SSB} problem is eliminated.  
Besides, parameters of feature representation of CVR network is shared with CTR network. The latter one is trained with much richer samples. This kind of parameter transfer learning~\cite{TLsurver} helps to alleviate the \textsl{DS} trouble remarkablely. 


For this work, we collect traffic logs from Taobao's recommender system. The full dataset consists of 8.9 billions samples with sequential labels of click and conversion.
Careful experiments are conducted. ESMM consistently outperforms competitive models, which demonstrate the effectiveness of the proposed approach. 
We also release our dataset\footnote[1]{https://tianchi.aliyun.com/datalab/dataSet.html?dataId=408\label{data_link}} for future research in this area.

\section{The proposed approach}\label{sec:opa}


\subsection{Notation} \label{sec:pf}
We assume the observed dataset to be $\mathcal{S} = \{(\bm{x}_i, y_i \rightarrow z_i)\}|_{i=1}^N$, with sample $(\bm{x}, y \rightarrow z)$ drawn from a distribution D with domain $\mathcal{X} \times \mathcal{Y} \times \mathcal{Z}$, where $\mathcal{X}$ is feature space, $\mathcal{Y}$ and $\mathcal{Z}$ are label spaces, and $N$ is the total number of impressions. 
$\bm{x}$ represents feature vector of observed impression, which is usually a high dimensional sparse vector with multi-fields~\cite{rendle2010factorization}, such as user field, item field etc. 
$y$ and $z$ are binary labels with $y=1$ or $z=1$ indicating whether click or conversion event occurs respectively.  
$y \rightarrow z$ reveals the sequential dependence of click and conversion labels that there is always a preceding click when conversion event occurs. 

Post-click CVR modeling is to estimate the probability of $\textit{pCVR} = p(z=1|y=1,\bm{x})$. Two associated probabilities are: post-view click-through rate (CTR) with $\textit{pCTR} = p(z=1|\bm{x})$ and post-view click\&conversion rate (CTCVR) with $\textit{pCTCVR} = p(y=1,z=1|\bm{x})$. Given impression $\bm{x}$, these probabilities follow Eq.(\ref{eq:pctcvr}):    
\begin{equation} \label{eq:pctcvr}
  \underbrace{p(y=1,z=1|\bm{x})}_{pCTCVR} 
= \underbrace{p(y=1|\bm{x}) }_{pCTR} \times 
  \underbrace{p(z=1|y=1,\bm{x})}_{pCVR}.
\end{equation}

\label{sec:ESMM}
\begin{figure}[!t]
\centering
\includegraphics[height=58mm ]{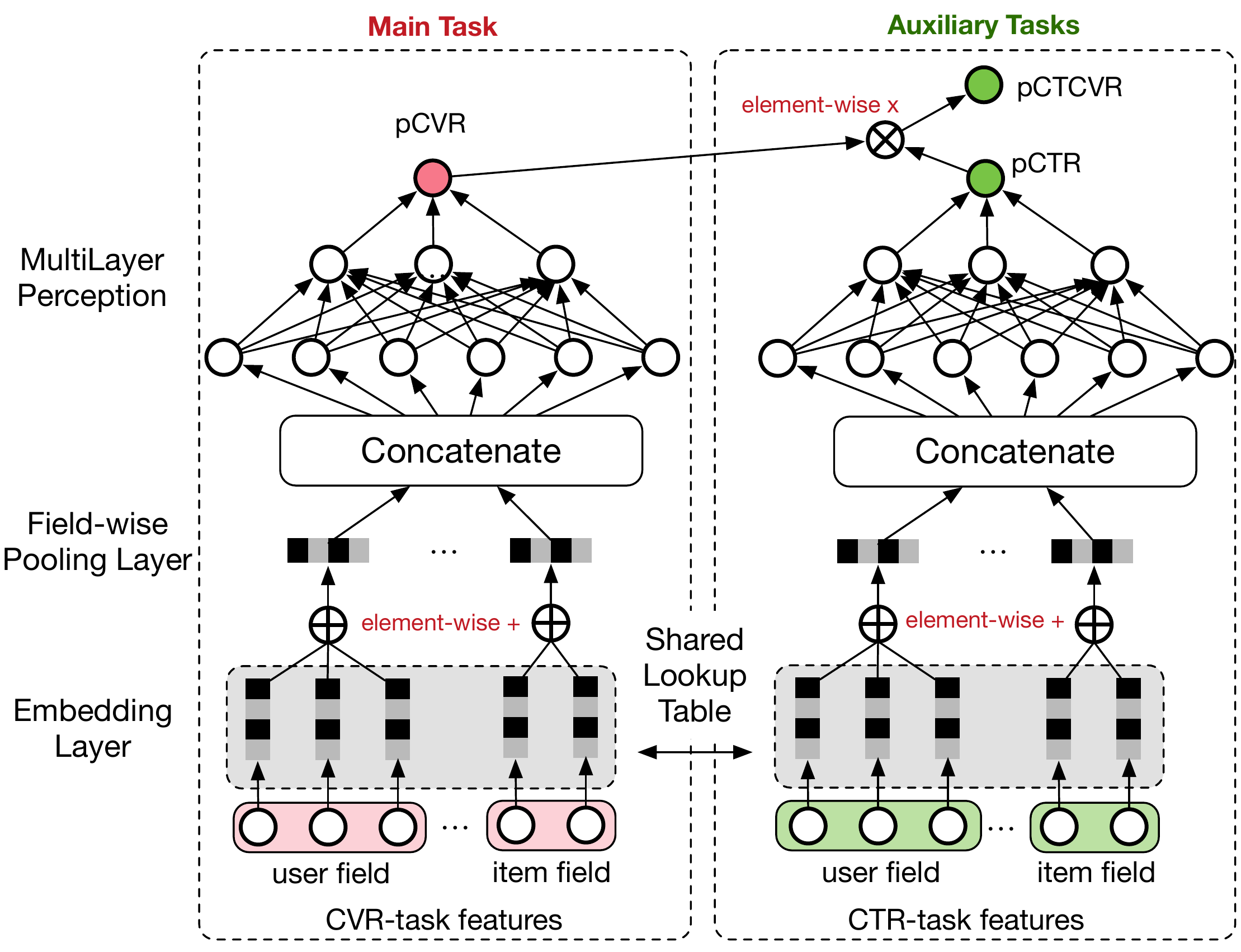}
\caption{{Architecture overview of ESMM for CVR modeling. In ESMM, two auxiliary tasks of CTR and CTCVR are introduced which: i) help to model CVR over entire input space, ii) provide feature representation transfer learning. ESMM mainly consists of two sub-networks: CVR network illustrated in the left part of this figure and CTR network in the right part. Embedding parameters of CTR and CVR network are shared. CTCVR takes the product of outputs from CTR and CVR network as the output. }}
\label{fig:mtl}
\end{figure}

\subsection{CVR Modeling and Challenges} \label{sec:baseline}

Recently deep learning based methods have been proposed for CVR modeling, achieving state-of-the-art performance.
Most of them follow a similar Embedding\&MLP network architecture, as introduced in \cite{din}. 
The left part of Fig.\ref{fig:mtl} illustrates this kind of architecture, which we refer to as \textbf{BASE} model, for the sake of simplicity.


In brief, conventional CVR modeling methods directly estimate the post-click conversion rate $p(z=1|y=1,\bm{x})$.
They train models with samples of clicked impressions, i.e., 
$ \mathcal{S}_c = \{(\bm{x}_j,  z_j) | y_j = 1\}|_{j=1}^{M} $. 
$M$ is the number of clicks over all impressions. 
Obviously, $\mathcal{S}_c$ is a subset of $ \mathcal{S}$. 
Note that in $\mathcal{S}_c$, (clicked) impressions without conversion are treated as negative samples and impressions with conversion (also clicked) as positive samples.   
In practice, CVR modeling encounters several task-specific problems, making it challenging.


\textbf{\textit{Sample selection bias (SSB)}}~\cite{zadrozny2004learning}.
In fact, conventional CVR modeling makes an approximation of $p(z=1|y=1,\bm{x}) \approx q(z=1|\bm{x}_c)$ by introducing an auxiliary feature space $\mathcal{X}_c$. $\mathcal{X}_c$ represents a \textsl{limited}\footnote[2]{space $\mathcal{X}_c$ equals to $\mathcal{X}$ under the condition that $\forall X \in \mathcal{X}, p(y=1|\bm{x}) > 0$ and the number of observed impressions is large enough. Otherwise, space $\mathcal{X}_c$ is part of $\mathcal{X}$.} space associated with $\mathcal{S}_c$. $\forall \bm{x}_c \in \mathcal{X}_c$ there exists a pair $(\bm{x}=\bm{x}_c,y_{\bm{x}}=1)$ where $\bm{x} \in \mathcal{X}$ and $y_{\bm{x}}$ is the click label of $\bm{x}$. In this way, $q(z=1|\bm{x}_c)$ is trained over space $\mathcal{X}_c$ with clicked samples of $\mathcal{S}_c$. At inference stage, the prediction of $p(z=1|y=1,\bm{x})$ over entire space $\mathcal{X}$ is calculated as $q(z=1|\bm{x})$ under the assumption that for any pair of $(\bm{x},y_{\bm{x}}=1)$ where $\bm{x} \in \mathcal{X}$, $\bm{x}$ belongs to $\mathcal{X}_c$. This assumption would be violated with a large probability as $\mathcal{X}_c$ is just a small part of entire space $\mathcal{X}$. It is affected heavily by the randomness of rarely occurred click event, whose probability varies over regions in space $\mathcal{X}$. Moreover, without enough observations in practice, space $\mathcal{X}_c$ may be quite different from $\mathcal{X}$. This would bring the drift of distribution of training samples from truly underling distribution and hurt the generalization performance for CVR modeling.

\textbf{\textit{Data sparsity (DS).}}
Conventional methods train CVR model with clicked samples of $\mathcal{S}_c$.
The rare occurrence of click event causes training data for CVR modeling to be extremely sparse.  
Intuitively, it is generally 1-3 orders of magnitude less than the associated CTR task, which is trained on dataset of $\mathcal{S}$ with all impressions. Table \ref{tab:dataset} shows the statistics of our experimental datasets, where number of samples for CVR task is just 4\% of that for CTR task.

It is worth mentioning that there exists other challenges for CVR modeling, e.g. \textit{delayed feedback}~\cite{delayConv}. 
This work does not focus on it. 
One reason is that the degree of conversion delay in our system is slightly acceptable. The other is that our approach can be combined with previous work~\cite{delayConv} to handle it.


\subsection{Entire Space Multi-Task Model}

The proposed \textbf{ESMM} is illustrated in Fig.\ref{fig:mtl}, which makes good use of the sequential pattern of user actions. 
Borrowing the idea from multi-task learning~\cite{ruder2017overview}, ESMM introduces two auxiliary tasks of CTR and CTCVR and eliminates the aforementioned problems for CVR modeling simultaneously.  

On the whole, ESMM simultaneously outputs $\textsl{pCTR}$, $\textsl{pCVR}$ as well as $\textsl{pCTCVR}$ w.r.t. a given impression. It mainly consists of two sub-networks: CVR network illustrated in the left part of Fig.\ref{fig:mtl} and CTR network in the right part. Both CVR and CTR networks adopt the same structure as \textbf{BASE} model. CTCVR takes the product of outputs from CVR and CTR network as the output.   
There are some highlights in ESMM, which have notable effects on CVR modeling and distinguish ESMM from conventional methods. 


\textbf{\textit{Modeling over entire space.}} Eq.(\ref{eq:pctcvr}) gives us hints, which can be transformed into Eq.(\ref{eq:cvr}).
\begin{equation} \label{eq:cvr}
\begin{split}
p(z=1|y=1,\bm{x}) = \frac{ p(y=1, z=1|\bm{x})} {p(y=1|\bm{x})}
\end{split}
\end{equation}
Here $p(y=1, z=1|\bm{x})$ and $p(y=1|\bm{x})$ are modeled on dataset of $\mathcal{S}$ with all impressions.
Eq.(\ref{eq:cvr}) tells us that with estimation of $\textsl{pCTCVR}$ and $\textsl{pCTR}$, $\textsl{pCVR}$ can be derived over the entire input space $\mathcal{X}$, which addresses the \textsl{sample selection bias} problem directly. 
This seems easy by estimating $\textsl{pCTR}$ and $\textsl{pCTCVR}$ with individually trained models separately and obtaining $\textsl{pCVR}$ by Eq.(\ref{eq:cvr}), which we refer to as \textbf{DIVISION} for simplicity.  
However, $\textsl{pCTR}$ is a small number practically, divided by which would arise numerical instability. 
ESMM avoids this with the multiplication form. 
In ESMM, $\textsl{pCVR}$ is just an intermediate variable which is constrained by the equation of Eq.(\ref{eq:pctcvr}). 
$\textit{pCTR}$ and $\textit{pCTCVR}$ are the main factors ESMM actually estimated over entire space. 
The multiplication form enables the three associated and co-trained estimators to exploit the sequential patten of data and communicate information with each other during training. Besides, it ensures the value of estimated $\textsl{pCVR}$ to be in range of [0,1], which in DIVISION method might exceed 1.   

The loss function of ESMM is defined as Eq.(\ref{loss:mtl}). It consists of two loss terms from CTR and CTCVR tasks which are calculated over samples of all impressions, without using the loss of CVR task. 

\begin{equation} \label{loss:mtl}
\begin{split} 
L(\theta_{cvr}, \theta_{ctr}) 
          &= \sum_{i=1}^N l\left(y_i, f(\bm{x}_i;\theta_{ctr})\right) \\ 
          &+ \sum_{i=1}^N l\left(y_i\&z_i, f(\bm{x}_i;\theta_{ctr}) \times f(\bm{x}_i;\theta_{cvr})\right),
\end{split}
\end{equation}
where $\theta_{ctr}$ and $\theta_{cvr}$ are the parameters of CTR and CVR networks and $l(\cdot)$ is cross-entropy loss function.  
Mathematically, Eq.(\ref{loss:mtl}) decomposes $y \rightarrow z$ into two parts\footnote[3]{Corresponding to labels of CTR and CTCVR tasks, which construct training datasets as follows: i) samples are composed of all impressions, ii) for CTR task, clicked impressions are labeled $y=1$, otherwise $y=0$, iii) for CTCVR task, impressions with click and conversion events occurred simultaneously are labeled $y\&z=1$, otherwise $y\&z=0$}: 
$y$ and $y\&z$, which in fact makes use of the sequential dependence of click and conversion labels. 
\textbf{\textit{Feature representation transfer.}}   
As introduced in section \ref{sec:baseline}, embedding layer maps large scale sparse inputs into low dimensional representation vectors. It contributes most of the parameters of deep network and learning of which needs huge volume of training samples.  
In ESMM, embedding dictionary of CVR network is shared with that of CTR network. It follows a feature representation transfer learning paradigm.
Training samples with all impressions for CTR task is relatively much richer than CVR task. 
This parameter sharing mechanism enables CVR network in ESMM to learn from un-clicked impressions and provides great help for alleviating the \textsl{data sparsity} trouble. 

Note that the sub-network in ESMM can be substituted with some recently developed models~\cite{widedeep,din}, which might get better performance. Due to limited space, we omit it and focus on tackling challenges encountered in real practice for CVR modeling.

\section{Experiments}
\subsection{Experimental Setup}
\textbf{\textit{Datasets.}} 
During our survey, no public datasets with sequential labels of click and conversion are found in CVR modeling area. 
To evaluate the proposed approach, we collect traffic logs from Taobao's recommender system and release a 1\% random sampling version of the whole dataset, whose size still reaches 38GB (without compression). 
In the rest of the paper, we refer to the released dataset as \textsl{\textbf{Public Dataset}} and the whole one as \textsl{\textbf{Product Dataset}}. 
Table \ref{tab:dataset} summarizes the statistics of the two datasets. 
Detailed descriptions can be found in the website of Public Dataset\textsuperscript{\ref{data_link}}.

\textbf{\textit{Competitors.}} We conduct experiments with several competitive methods on CVR modeling. 
(1) \textbf{\uppercase{Base}} is the baseline model introduced in section \ref{sec:baseline}.      
(2) \textbf{AMAN}~\cite{pan2008one} applies negative sampling strategy and best results are reported with sampling rate searched in $\{10\%, 20\%, 50\%, 100\%\}$.  
(3) \textbf{\uppercase{Oversampling}}~\cite{weiss2004mining} copies positive examples to reduce difficulty of training with sparse data, with sampling rate searched in $\{2,3,5,10\}$. 
(4) \textbf{\uppercase{Unbias}} follows \cite{zhang2016bid} to fit the truly underlying distribution from observations via rejection sampling. $\textit{pCTR}$ is taken as the rejection probability. 
(5) \textbf{\uppercase{Division}} estimates $\textit{pCTR}$  and $\textit{pCTCVR}$ with individually trained CTR and CTCVR networks and calculates $\textit{pCVR}$ by Eq.(\ref{eq:cvr}).
(6) \textbf{\uppercase{ESMM-ns}} is a lite version of ESMM without sharing of embedding parameters.

The first four methods are different variations to model CVR directly based on state-of-the-art deep network. 
DIVISION, ESMM-NS and ESMM share the same idea to model CVR over entire space which involve three networks of CVR, CTR and CTCVR.
ESMM-NS and ESMM co-train the three networks and take the output from CVR network for model comparison.             
To be fair, all competitors including \uppercase{ESMM} share the same  network structure and hyper parameters with \uppercase{Base} model, which i) uses ReLU activation function, ii) sets the dimension of embedding vector to be 18, iii) sets dimensions of each layers in MLP network to be $360 \times 200 \times 80 \times 2$, iv) uses adam solver with parameter \textbf{$\beta_1 = 0.9,\beta_2 = 0.999,\epsilon = 10^{-8}$}. 


\textbf{\textit{Metric.}} The comparisons are made on two different tasks: (1) conventional CVR prediction task which estimates \textsl{pCVR} on dataset with clicked impressions, (2) CTCVR prediction task which estimates \textsl{pCTCVR} on dataset with all impressions. 
Task (2) aims to compare different CVR modeling methods over entire input space, which reflects the model performance corresponding to \textsl{SSB} problem. 
In CTCVR task, all models calculate $\textsl{pCTCVR}$ by $\textsl{pCTR} \times \textsl{pCVR}$, where: 
i) $\textsl{pCVR}$ is estimated by each model respectively,  
ii) $\textsl{pCTR}$ is estimated with a same independently trained CTR network (same structure and hyper parameters as BASE model).
Both of the two tasks split the first 1/2 data in the time sequence to be training set while the rest to be test set.       
Area under the ROC curve (AUC) is adopted as performance metrics.  
All experiments are repeated 10 times and averaged results are reported.

\begin{table}[!t]
\caption{{Statistics of experimental datasets.}}
\label{tab:dataset}
\small
\begin{threeparttable}
\centering{
 \centerline {\resizebox{0.47\textwidth}{!}{
\begin{tabular}{lcccccc}
\toprule
dataset & \#user & \#item & \#impression & \#click & \#conversion \\
\midrule
Public Dataset & 0.4M & 4.3M & 84M & 3.4M & 18k \\
Product Dataset & 48M &23.5M & 8950M & 324M & 1774k \\
\bottomrule
\end{tabular}}}}
\end{threeparttable}
\end{table}

\begin{table}[]
\caption{Comparison of different models on Public Dataset.}
\label{tab:public}
\small
\centering{
 \centerline {\resizebox{0.45\textwidth}{!}{
 \begin{tabular}{lcc}
   \toprule
    Model       & AUC(mean \textpm\ std) on CVR task & AUC(mean \textpm\ std) on CTCVR task \\ \midrule
   \uppercase{Base}          & 66.00 \textpm\ 0.37           & 62.07 \textpm\ 0.45  \\
   \uppercase{AMAN}          & 65.21 \textpm\ 0.59           & 63.53 \textpm\ 0.57 \\
   \uppercase{Oversampling}  & 67.18 \textpm\ 0.32           & 63.05 \textpm\ 0.48 \\
   \uppercase{Unbias}        & 66.65 \textpm\ 0.28           & 63.56 \textpm\ 0.70  \\
   \uppercase{Division}      & 67.56 \textpm\ 0.48           & 63.62 \textpm\ 0.09  \\
   \textbf{ESMM-NS}          & \textbf{68.25 \textpm\ 0.44}  & \textbf{64.44 \textpm\ 0.62} \\
   \textbf{ESMM}             & \textbf{68.56 \textpm\ 0.37}  & \textbf{65.32 \textpm\ 0.49} \\
   \bottomrule
\end{tabular}}}}
\end{table}

\subsection{Results on Public Dataset}
Table \ref{tab:public} shows results of different models on public dataset.
(1) Among all the three variations of \uppercase{Base} model, only \uppercase{AMAN} performs a little worse on CVR task, which may be due to the sensitive of random sampling. \uppercase{Oversampling} and \uppercase{Unbias} show improvement over \uppercase{Base} model on both CVR and CTCVR tasks. 
(2) Both \uppercase{Division} and ESMM-NS estimate $\textsl{pCVR}$ over entire space and achieve remarkable promotions over \uppercase{Base} model. Due to the avoidance of numerical instability, ESMM-NS performs better than \uppercase{Division}.
(3) ESMM further improves ESMM-NS. 
By exploiting the sequential patten of user actions and learning from un-clicked data with transfer mechanism, ESMM provides an elegant solution for CVR modeling to eliminate \textsl{SSB} and \textsl{DS} problems simultaneously and beats all the competitors.  
Compared with BASE model, ESMM achieves absolute AUC gain of 2.56\% on CVR task, which indicates its good generalization performance even for biased samples. 
On CTCVR task with full samples, it brings 3.25\% AUC gain.
These results validate the effectiveness of our modeling method. 

\subsection{Results on Product Dataset}
\begin{figure}[!t]
\centering
\includegraphics[height=37mm ]{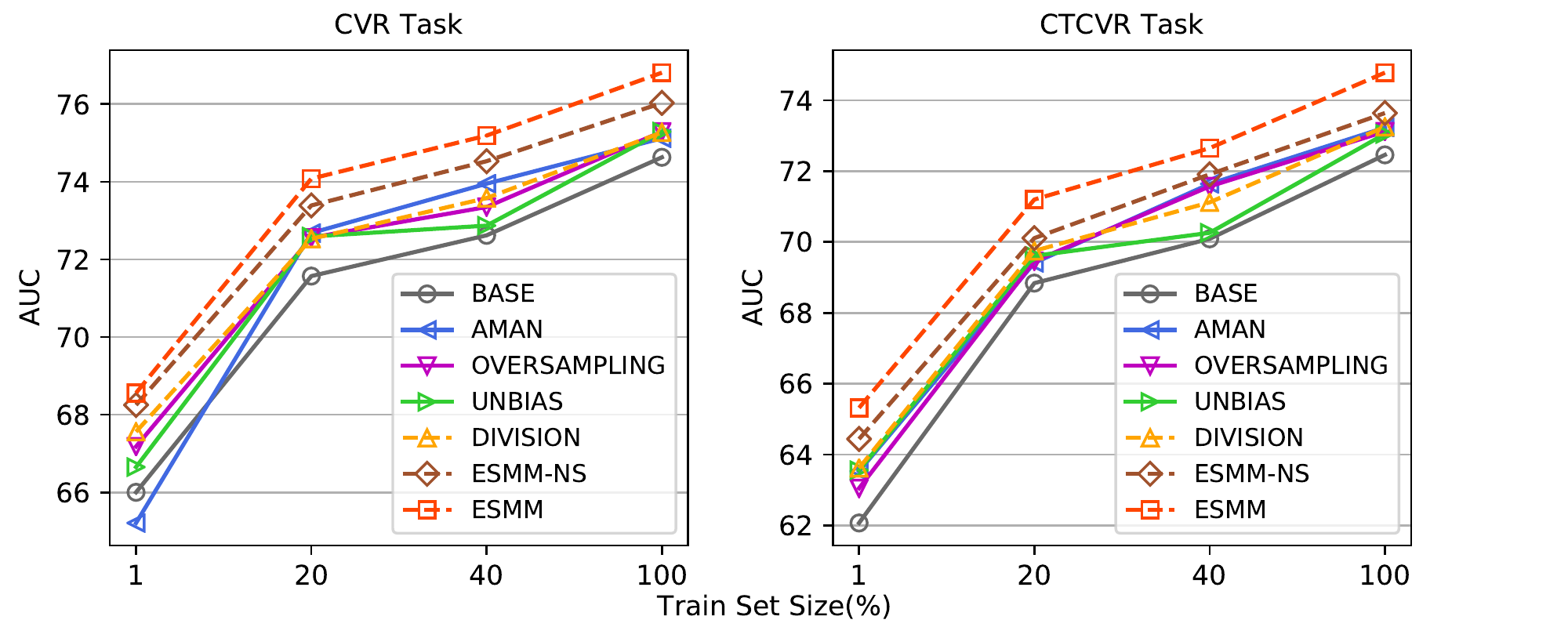}
\caption{{Comparison of different models w.r.t. different sampling rates on Product Dataset.}}
\label{fig:productive}
\end{figure}

We further evaluate ESMM on our product dataset with 8.9 billions of samples, two orders of magnitude larger than public one. 
To verify the impact of the volume of the training dataset, we conduct careful comparisons on this large scale datasets w.r.t. different sampling rates, as illustrated in Fig.\ref{fig:productive}. 
First, all methods show improvement with the growth of volume of training samples. This indicates the influence of data sparsity. In all cases except AMAN on 1\% sampling CVR task, BASE model is defeated.  
Second, ESMM-NS and ESMM outperform all competitors consistently w.r.t. different sampling rates. 
In particular, ESMM maintains a large margin of AUC promotion over all competitors on both CVR and CTCVR tasks.
BASE model is the latest version which serves the main traffic in our real system. 
Trained with the whole dataset, ESMM achieves absolute AUC gain of 2.18\% on CVR task and 2.32\% on CTCVR task over BASE model.
This is a significant improvement for industrial applications where 0.1\% AUC gain is remarkable.

\vspace{-0.1cm}

\section{Conclusions and Future Work}
In this paper, we propose a novel approach ESMM for CVR modeling task. 
ESMM makes good use of sequential patten of user actions. 
With the help of two auxiliary tasks of CTR and CTCVR, ESMM elegantly tackles challenges of \textit{sample selection bias} and \textit{data sparsity} for CVR modeling encountered in real practice. 
Experiments on real dataset demonstrate the superior performance of the proposed ESMM.   
This method can be easily generalized to user action prediction in scenario with sequential dependence.   
In the future, we intend to design global optimization models in applications with multi-stage actions like $request \rightarrow impression \rightarrow click \rightarrow conversion$.     


\bibliographystyle{ACM-Reference-Format}
\bibliography{reference}

\end{document}